\documentclass{firered}
\usepackage[utf8]{inputenc} 
\usepackage[T1]{fontenc}    
\usepackage{hyperref}       
\usepackage{url}            
\usepackage{booktabs}       
\usepackage{amsfonts}       
\usepackage{nicefrac}       
\usepackage{microtype}      
\usepackage{float}
\usepackage{mwe}
\usepackage{graphicx}
\usepackage{amssymb}
\usepackage{adjustbox}
\usepackage{colortbl}
\usepackage{array}
\usepackage{multirow}
\usepackage{makecell}
\usepackage{bbm}
\usepackage{collcell,xfp}
\usepackage{pgf}
\usepackage{tikz}
\usepackage[most]{tcolorbox}
\usepackage{csquotes}
\usepackage[noorphans,vskip=1em,leftmargin=1em]{quoting}
\usepackage{pifont} 
\usepackage{enumitem} 
\usepackage{tcolorbox} 
\usepackage{forest}
\usepackage{wrapfig}
\usepackage{caption}
\usepackage{subcaption}
\usepackage{longtable}
\usepackage{algpseudocode}
\usepackage{amsmath}
\usepackage[capitalize]{cleveref}
\usepackage[ruled,vlined,linesnumbered]{algorithm2e} 
\usepackage[percent]{overpic}
\usepackage{xspace}
\usepackage[normalem]{ulem}  
\usepackage{tocloft}  
\usepackage{fancyhdr} 
\usepackage{natbib}  

\fancypagestyle{firststyle}{
    \fancyhead[L]{}
    \fancyhead[C]{}
    \fancyhead[R]{}
    
    \renewcommand{\footrulewidth}{0.8pt}
    \renewcommand{\headrule}{}
    \renewcommand{\footrule}{\color{sectionred}\hrule width\textwidth height\footrulewidth \vspace{-100pt}}
    \fancyfoot[C]{}
}

\renewcommand{\headrule}{}
\renewcommand{\footrulewidth}{0pt}

\usepackage{listings}
\usepackage{wrapfig}
\usepackage{geometry}
\usepackage{xcolor}
\usepackage[most]{tcolorbox}

\usepackage{CJKutf8}   

\usepackage{arydshln}
\usepackage{twemojis}
\usepackage{tikz-dependency}
\usepackage{graphicx}
\usepackage{tikz}
\usepackage{tikz-qtree}
\usepackage{caption}
\usetikzlibrary{arrows.meta}
\definecolor{clearpurple}{RGB}{138, 140, 191}
\definecolor{clearyellow}{HTML}{f2d3bf}
\definecolor{skyblue}{HTML}{a8d8e2}
\definecolor{darkblue}{HTML}{19183B}
\definecolor{tp1}{RGB}{253, 207, 158}

\usepackage{fontawesome}
\newcommand{\huggingface}{\raisebox{-1.5pt}{\includegraphics[height=1.05em]{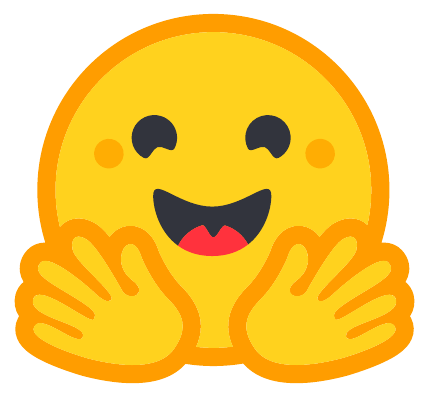}}\xspace}

\newcommand{\github}{\raisebox{-1.5pt}{\includegraphics[height=1.05em]{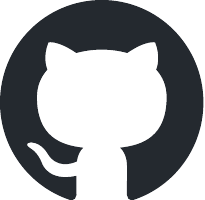}}\xspace}

\usepackage[most]{tcolorbox} 

\newtcolorbox{insight}[1][]{
  colback=red!5!white, 
  colframe=red!75!black, 
  fonttitle=\bfseries, 
  title=Insight, 
  sharp corners=downhill, 
  enhanced, 
  attach boxed title to top left={yshift=-2mm, xshift=2mm}, 
  boxed title style={colback=red!75!black}, 
  #1
}

\definecolor{scholarblue}{rgb}{0.21,0.49,0.74}
\definecolor{bluelink}{RGB}{0,113,188}
\definecolor{greenlink}{RGB}{0,188,113}
\hypersetup{
    colorlinks=true,%
    citecolor=scholarblue,%
    filecolor=red,%
    linkcolor=red!93!black,%
    urlcolor=bluelink
}

\newtcolorbox{promptbox}[1][]{
    enhanced,
    colback=gray!10,      
    colframe=black,       
    coltitle=white,       
    fonttitle=\bfseries\sffamily, 
    fontupper=\ttfamily\small,    
    boxrule=0.8pt,        
    arc=4mm,              
    title={#1},           
    attach boxed title to top left={xshift=0.5cm, yshift*=-\tcboxedtitleheight/2},
    boxed title style={
        colback=black,    
        arc=3mm,          
        boxrule=0pt,      
        left=3mm, right=3mm, top=1.5mm, bottom=1.5mm
    },
    top=1.5em,            
    parbox=false          
}

\usepackage{titlesec}
\titlespacing*{\paragraph}
    {0pt}     
    {0.25em}  
    {1em}  

\definecolor{sectionred}{RGB}{220,38,38}  
\titleformat{\section}{\color{sectionred}\large\bfseries}{\color{sectionred}\thesection.}{0.5em}{#1}[]
\titleformat{\subsection}{\color{sectionred}\bfseries}{\color{sectionred}\thesubsection.}{0.5em}{#1}[]
\titleformat{\subsubsection}{\color{sectionred}\bfseries\itshape}{\color{sectionred}\thesubsubsection.}{0.5em}{#1}[]

\definecolor{cGreen}{RGB}{69, 139, 89}    
\definecolor{cYellow}{RGB}{241, 186, 88}  
\definecolor{cBlue}{RGB}{95, 126, 222}    
\definecolor{cRed}{RGB}{219, 86, 96}      
\definecolor{cPurple}{RGB}{166, 156, 226} 
\definecolor{gridColor}{RGB}{200, 200, 200}
\definecolor{bgColor}{RGB}{248, 252, 252}

\definecolor{navyblue}{HTML}{0071BC}
\iftrue   
\newcommand{\displaytodo}[1]{#1}
\else
\newcommand{\displaytodo}[1]{}
\fi

\definecolor{blindcolor}{HTML}{AB2AC6}    
\definecolor{chancecolor}{HTML}{F59E0B}   
\definecolor{singlecolor}{HTML}{06B6D4}   
\definecolor{multiplecolor}{HTML}{2563EB} 
\definecolor{captioncolor}{HTML}{22C55E}  

 \newcommand{\culine}[2]{%
    \def\temp@uline{\bgroup\markoverwith
        {\textcolor{#1}{\rule[-0.5ex]{2pt}{1pt}}}\ULon}%
    \temp@uline{#2}%
}
 \newcommand{\cthickuline}[3][0.8pt]{%
    \def\temp@uline{\bgroup\markoverwith
        {\textcolor{#2}{\rule[-0.5ex]{2pt}{#1}}}\ULon}%
    \temp@uline{#3}%
}

\title{\center{FireRed-OCR Technical Report}} 

\renewcommand\Affilfont{\normalfont\fontsize{11}{15}\selectfont\centering}
\author{ 
    Super Intelligence Team, Xiaohongshu Inc.
}

\begin{document}

{\color{sectionred}\hrule width\textwidth height0.2pt}

\renewcommand{\abstractname}{\textcolor{sectionred}{Abstract}}
\begin{abstract}
We present FireRed-OCR, a systematic framework to specialize general VLMs into high-performance OCR models. Large Vision-Language Models (VLMs) have demonstrated impressive general capabilities but frequently suffer from "structural hallucination" when processing complex documents, limiting their utility in industrial OCR applications. In this paper, we introduce FireRed-OCR, a novel framework designed to transform general-purpose VLMs (based on Qwen3-VL) into pixel-precise structural document parsing experts. To address the scarcity of high-quality structured data, we construct a "Geometry + Semantics" Data Factory. Unlike traditional random sampling, our pipeline leverages geometric feature clustering and multi-dimensional tagging to synthesize and curate a highly balanced dataset, effectively handling long-tail layouts and rare document types. Furthermore, we propose a Three-Stage Progressive Training strategy that guides the model from pixel-level perception to logical structure generation. This curriculum includes: (1) Multi-task Pre-alignment to ground the model’s understanding of document structure; (2) Specialized SFT for standardizing full-image Markdown output; and (3) Format-Constrained Group Relative Policy Optimization (GRPO), which utilizes reinforcement learning to enforce strict syntactic validity and structural integrity (e.g., table closure, formula syntax). Extensive evaluations on OmniDocBench v1.5 demonstrate that FireRed-OCR achieves state-of-the-art performance with an overall score of 92.94\%, significantly outperforming strong baselines such as DeepSeek-OCR 2 and OCRVerse across text, formula, table, and reading order metrics. We open-source our code and model weights to facilitate the "General VLM to Specialized Structural Expert" paradigm.

\end{abstract}

\begin{figure}[b]
  \centering
  \includegraphics[width=0.9\textwidth]{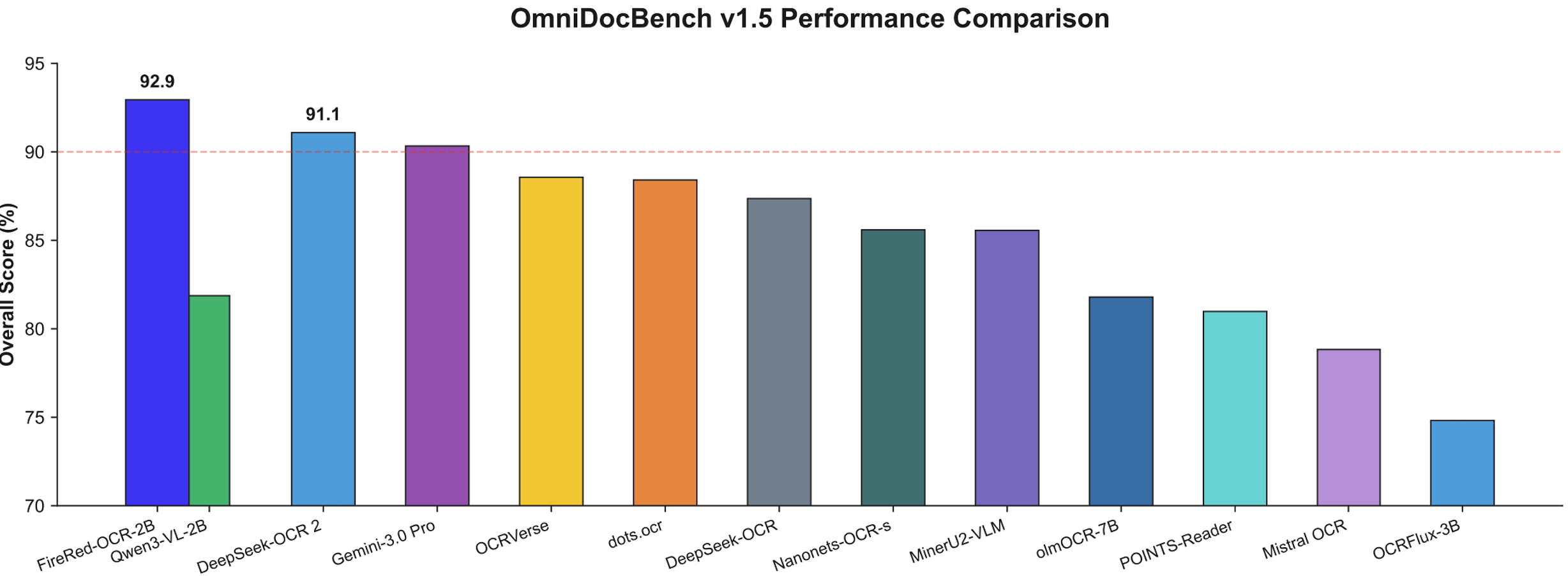}
  \caption{Performance comparison on the OmniDocBench v1.5 benchmark. FireRed-OCR achieves state-of-the-art performance, securing the top rank in the overall evaluation with a score exceeding 90\%.}
  \label{fig:introduction_benchmark}
\end{figure}



\setlength{\parindent}{0pt}

\begin{tikzpicture}[remember picture,overlay]
            \node[anchor=north west, xshift=1.5cm, yshift=-1.0cm] at (current page.north west) {
                \includegraphics[width=5.4cm]{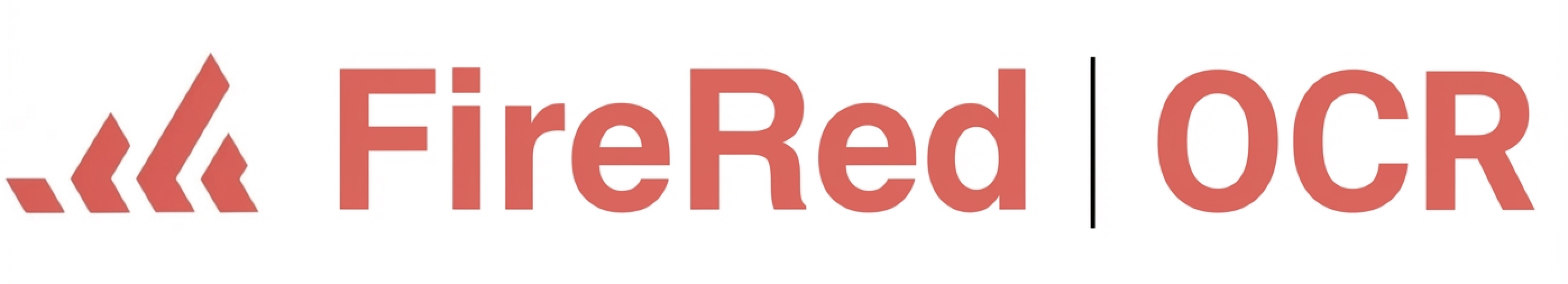}
            };
\end{tikzpicture}

\maketitle

\begin{center}
    \renewcommand{\arraystretch}{1.5}
    \begin{tabular}{rll}

        \github{} & \textbf{GitHub} & {\textcolor{sectionred}{\href{https://github.com/FireRedTeam/FireRed-OCR}{https://github.com/FireRedTeam/FireRed-OCR}}} \\
        \huggingface{} & \textbf{HuggingFace Model} & {\textcolor{sectionred}{\href{https://huggingface.co/FireRedTeam/FireRed-OCR}{https://huggingface.co/FireRedTeam/FireRed-OCR}}} \\
        \huggingface{} & \textbf{HuggingFace Demo} & {\textcolor{sectionred}{\href{https://huggingface.co/spaces/FireRedTeam/FireRed-OCR}{Online Demo (HuggingFace)}}} \\
    \end{tabular}
\end{center}

\setcounter{footnote}{0}  
\renewcommand{\thefootnote}{\arabic{footnote}}  

\clearpage
\renewcommand{\contentsname}{\textcolor{sectionred}{Contents}}
\tableofcontents
\clearpage

\section{Introduction}
\begin{figure}[t]
  \centering
  \includegraphics[width=0.99\textwidth]{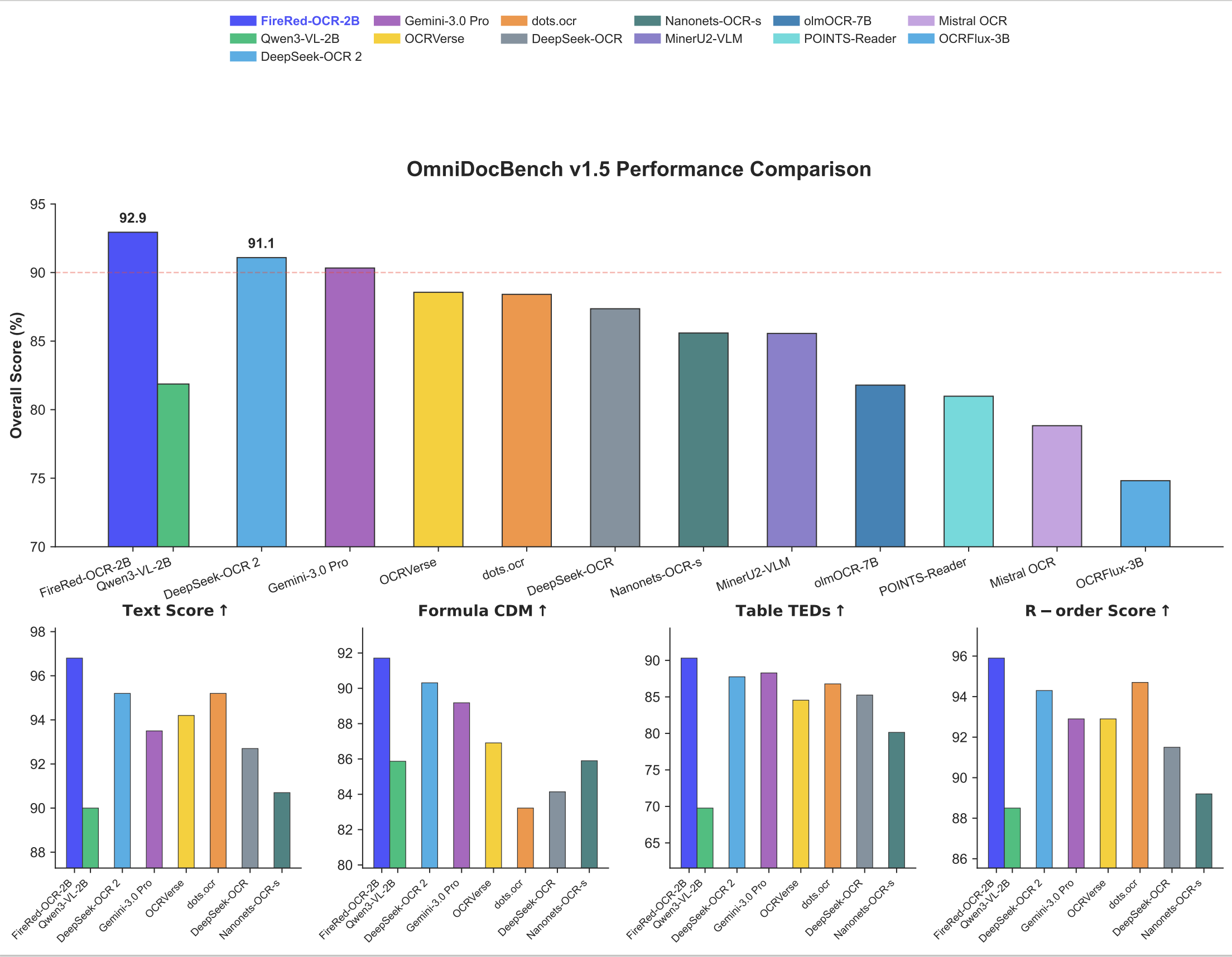}
  \caption{Comparison with state-of-the-art models on OmniDocBench v1.5~\cite{ouyang2025omnidocbench}. FireRed-OCR ranks first in the overall metric (top) and demonstrates leading performance across all sub-categories: Text recognition, Formula processing, Table structure analysis, and Reading order inference (bottom).}
  \label{fig:introduction_benchmark}
\end{figure}
The rapid evolution of Large Vision-Language Models (VLMs)~\cite{chen2023vlp,wang2024qwen2,chen2023x,zhang2024mm} has marked a transformative milestone in multimodal artificial intelligence~\cite{zhang2023structure,google2025gemini3pro}. Models such as GPT-4V~\cite{hurst2024gpt}, Qwen-VL~\cite{bai2023qwen}, and LLaVA~\cite{liu2023visual} have demonstrated remarkable capabilities in general image captioning and reasoning. However, these general-purpose models often exhibit significant instability when applied to Document Intelligence tasks involving the parsing of financial reports, academic papers, and complex forms. While they possess the semantic capacity to interpret visual content, they frequently fail to adhere to rigorous formatting constraints. We define this phenomenon as \textbf{Structural Hallucination}, which manifests as disordered rows in Markdown tables, non-existent syntax in mathematical formulas, or the omission of hierarchical logic. Such errors render the outputs effectively unusable for downstream industrial applications. The fundamental challenge, therefore, lies in transitioning the model's behavior from broad semantic interpretation to precise structural generation governed by strict logical rules.

Traditional OCR systems~\cite{cui2025paddleocr,cui2026paddleocr,zhang2025monkeyocr}, such as PaddleOCR-VL~\cite{cui2025paddleocr}, rely on pipeline approaches (detection followed by recognition). While pixel-precise, they often lack the semantic understanding required to reconstruct logical structures like reading order in multi-column layouts. Conversely, recent End-to-End approaches, including DeepSeek-OCR~\cite{wei2025deepseek}, DeepSeek-OCR 2~\cite{wei2026deepseek}, OCRVerse~\cite{zhong2026ocrverse}, and varying generic VLMs~\cite{bai2025qwen3,chen2025mindvl,team2023gemini,google2025gemini3pro}, treat document parsing as a sequence generation task. Although these models capture semantics better than traditional pipelines, they struggle with fine-grained spatial grounding. Existing general VLMs, despite their size, often lack the specialized alignment data and reinforcement mechanisms required to suppress hallucinations in dense text environments. They understand the ``intent'' but not the ``rules.''

To bridge this gap, we present \textbf{FireRed-OCR}, a framework designed to evolve a general VLM (specifically Qwen3-VL~\cite{bai2025qwen3}) into an industrial-grade structural expert. Our approach is founded on two pillars: a high-precision data engine and a progressive training strategy.

First, we address the data quality bottleneck by constructing a \textbf{``Geometry + Semantics'' Data Factory}. Traditional random sampling fails to capture the diversity of document layouts. We introduce a dual-indexing mechanism that clusters documents based on geometric features and semantic tags, enabling balanced sampling of rare layouts. We further implement an automated synthesis pipeline for scarce data and employ a ``Expert-Level Refinement'' process using advanced models to correct hard cases.

Second, we propose a \textbf{Three-Stage Progressive Training} pipeline to ``tame'' the model:
\begin{enumerate}
    \item \textbf{Multi-task Pre-alignment:} We force the model to learn physical perception (bounding boxes and region recognition) before attempting full-page generation.
    \item \textbf{Specialized SFT:} We refine the model on high-quality Markdown data to ensure standard syntax adherence.
    \item \textbf{Format-Constrained GRPO:} We introduce a Reinforcement Learning stage using Group Relative Policy Optimization (GRPO)~\cite{shao2024deepseekmath} with specific rewards for formula validity, table closure, and content accuracy, endowing the model with self-correction capabilities.
\end{enumerate}

Our contributions are summarized as follows:
\begin{itemize}
    \item We identify \textit{Structural Hallucination} as the primary bottleneck in VLM-based document parsing and propose a solution that shifts the paradigm from text generation to structural engineering.
    \item We develop a ``Geometry + Semantics'' Data Factory that ensures interpretable, balanced, and high-quality supervision data.
    \item We introduce a three-stage training strategy incorporating constrained GRPO, significantly improving robustness in complex layout reconstruction.
    \item FireRed-OCR achieves SOTA results on the authoritative \textbf{OmniDocBench v1.5}, surpassing DeepSeek-OCR2 and OCRVerse with a 92.94\% overall score, proving the viability of our ``General VLM $\rightarrow$ Specialized Structural Model'' paradigm.
\end{itemize}

\section{Data}
\begin{figure}[h]
  \centering
  \includegraphics[width=0.99\textwidth]{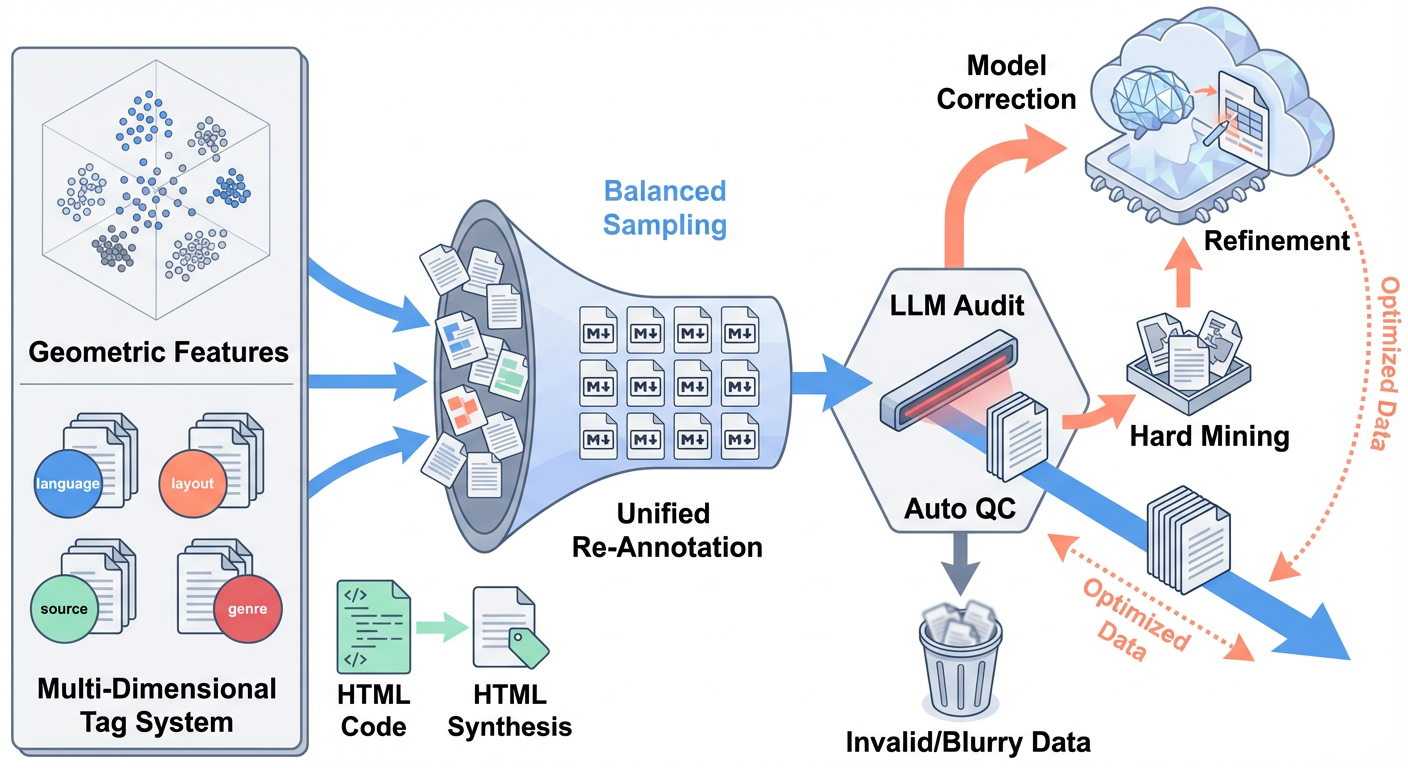}
  \caption{Overview of the proposed data processing and optimization pipeline. This pipeline operates through five rigorous stages, including Geometry-Driven Feature Extraction and Dual Indexing, Stratified Sampling and Unified Re-annotation, Synthetic Data Generation for Structural Priors, Automated Quality Control and Hard-Negative Mining and Expert-Level Refinement via Model Distillation.}
  \label{fig:data}
\end{figure}
The performance ceiling of structure-aware VLMs is determined not by the sheer volume of data, but by the density of high-quality, structurally diverse samples. Traditional OCR datasets~\cite{chng2019icdar2019,singh2021textocr} often suffer from two critical issues: (1) \textit{Distribution Imbalance}, where simple layouts (e.g., novels) dominate over complex ones (e.g., multi-column papers, financial tables), and (2) \textit{Annotation Inconsistency}, where different sources utilize conflicting Markdown standards.

To address these challenges, we introduce the \textbf{``Geometry + Semantics'' Data Factory}, an automated, scalable refinery that transforms raw web data into ``high-octane fuel'' for structural learning. This pipeline operates through five rigorous stages.

\subsection{Geometry-Driven Feature Extraction and Dual Indexing}
Random sampling is insufficient for ensuring layout diversity. A text-heavy page and a table-heavy page may share similar semantic topics but require vastly different structural parsing capabilities. We propose a \textbf{Dual-Indexing Mechanism} to decouple layout complexity from semantic content.

\subsubsection{Geometric Clustering}
We utilize a lightweight, pre-trained image encoder (e.g., ResNet/ViT variants)~\cite{radford2021learning,chuang2025meta} to extract visual feature vectors from document images, deliberately ignoring textual content to focus on \textit{layout topology}. By performing unsupervised clustering (e.g., K-Means~\cite{ahmed2020k} or DBSCAN~\cite{hahsler2019dbscan}) on these geometric embeddings, we map the dataset into a structural feature space.
\begin{itemize}
    \item \textbf{Redundancy Elimination:} We identify and downsample clusters with high density (e.g., standard single-column text), removing highly similar duplicates.
    \item \textbf{Rare Layout Mining:} We identify sparse clusters representing ``long-tail'' layouts, such as nested tables, irregular forms, or artistic typography, ensuring these hard examples are preserved.
\end{itemize}

\subsubsection{Multi-dimensional Tagging System}
Complementary to geometric features, we construct a semantic tag system to ensure interpretability. We employ a classifier to label data across four dimensions:
\begin{itemize}
    \item \textbf{Language:} Balancing dominant languages with low-resource ones.
    \item \textbf{Layout Type:} Dense text, multi-column, table-heavy, image-rich.
    \item \textbf{Document Source:} Scanned PDFs, camera-captured images, digital-born documents.
    \item \textbf{Genre:} Academic papers, financial invoices, legal contracts, receipts.
\end{itemize}
This dual-index allows for precise control over the training distribution, enabling us to curate a dataset that is both geometrically diverse and semantically balanced.

\subsection{Stratified Sampling and Unified Re-annotation}
Existing open-source datasets~\cite{singh2021textocr,olmocrbench} often possess conflicting Ground Truth (GT) styles. Some represent formulas with ``$$...$$'', others with ``\(...\)''; some use HTML tables, others Markdown. Such inconsistency confuses the model during convergence.

\subsubsection{Balanced Sampling}
Leveraging the dual-index, we perform stratified sampling. We heavily upsample the ``rare layout'' clusters and ``complex genre'' tags while capping the contribution of simple text documents. This ensures that the model sees a uniform distribution of \textit{difficulty} rather than a natural distribution of data.

\subsubsection{Standardized Markdown Re-annotation}
To eliminate style conflicts, we discard original disparate annotations and re-process the entire sampled corpus using \textbf{PaddleOCR-VL}~\cite{cui2025paddleocr}. While PaddleOCR-VL serves as a strong baseline, we utilize it primarily to standardize the \textit{format} (unified Markdown syntax). This step creates a cohesive ``lingua franca'' for the model, ensuring that `Level 1 Headers` or `Bold Text` are represented consistently across 10 million+ samples, regardless of their original source.

\subsection{Synthetic Data Generation for Structural Priors}
Despite aggressive mining, certain structural combinations (e.g., tables spanning three pages, complex nested mathematical proofs) remain scarce in natural data. To bridge this gap, we implement a \textbf{Render-based Synthesis Pipeline}.
We construct a library of HTML/CSS templates capable of generating complex layouts procedurally. By randomizing content (text, numbers, formulas) and style parameters (border visibility, column spans, font types), we render high-resolution images with perfectly aligned Markdown Ground Truths. This provides the model with error-free structural priors, particularly for:
\begin{itemize}
    \item \textbf{Complex Tables:} Spanning cells, invisible borders, and color-coded rows.
    \item \textbf{Mathematical Formulas:} Heavy nesting and rare symbols.
\end{itemize}

\subsection{Automated Quality Control and Hard-Negative Mining}
Quality control is critical to preventing ``Garbage In, Garbage Out.'' We implement a bi-level filtration system designed not just to clean data, but to identify valuable ``hard negatives.''

\subsubsection{Automated Heuristics (The Sieve)}
We apply rule-based checks to the Markdown ground-truth:
\begin{itemize}
    \item \textbf{Structural Closure:} Verifying that all opened tags (e.g., `table`, `**`) are correctly closed.
    \item \textbf{Table Integrity:} Checking that the number of Markdown pipe characters `|` matches the expected column count for every row.
    \item \textbf{Garbage Detection:} Filtering out samples with high ratios of non-printable characters or excessive repetition.
\end{itemize}

\subsubsection{LLM-based Content Audit (The Judge)}
Data passing the heuristic filter is subjected to a lightweight LLM audit. The LLM classifies potential issues into:
\begin{itemize}
    \item \textbf{Discard (Noise):} Blank pages, severe blur, substantial content cut-off.
    \item \textbf{Keep (Hard Case):} Samples where the content is legible but the layout is highly complex or the initial OCR failed.
\end{itemize}
Crucially, we do not discard the ``Hard Cases.'' Instead, they are routed to a specialized repository for refinement.

\subsection{Expert-Level Refinement via Model Distillation}
The \textit{HardCase} repository represents the "frontier" of the model's capability. To unlock the value in these samples, we employ an expert-in-the-loop strategy using advanced proprietary models (e.g., Gemini 3 Pro series~\cite{google2025gemini3pro}).
We feed the HardCase images to these teacher models with a prompt specifically designed to correct structural errors while preserving the standardized Markdown format. This process:
\begin{enumerate}
    \item \textbf{Repairs Structural Hallucinations:} Fixing missing rows in tables or disordered text blocks in the GT.
    \item \textbf{Reduces Systemic Bias:} Mitigating errors introduced by the initial PaddleOCR pass.
\end{enumerate}
This ``Expert Consultation'' effectively distills the reasoning power of massive frontier models into our training set, providing FireRed-OCR with pixel-perfect supervision for the most challenging document scenarios.

\section{Method}
\begin{figure}[h]
  \centering
  \includegraphics[width=0.99\textwidth]{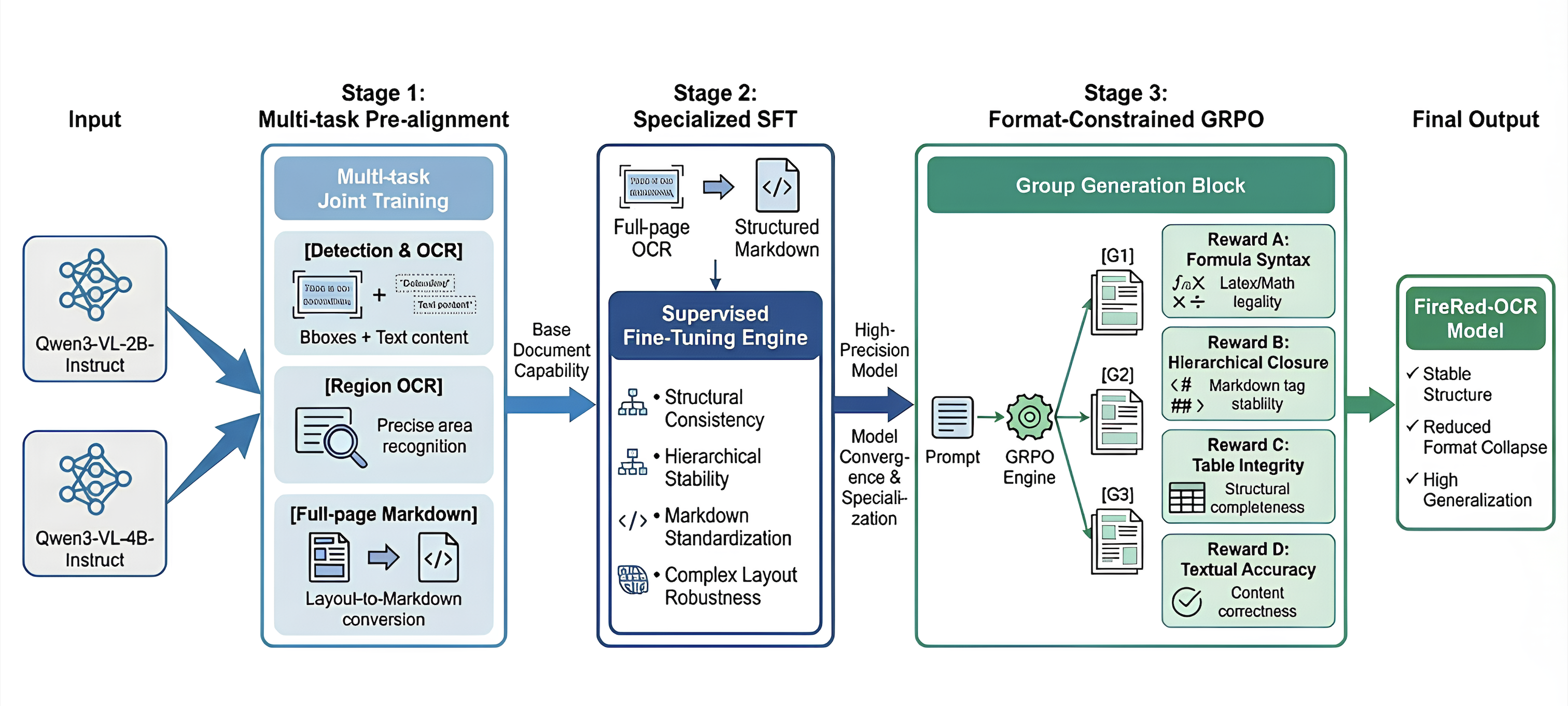}
  \caption{Overview of the FireRed-OCR training framework. The pipeline progresses through three stages: (1) Multi-task Pre-alignment, which employs joint training on detection, region OCR, and layout tasks to ground the model’s understanding of document structure; (2) Specialized Supervised Fine-Tuning (SFT), designed to adapt base capabilities for end-to-end structured Markdown generation; and (3) Format-Constrained GRPO, a reinforcement learning stage incorporating rule-based rewards (targeting formula syntax, hierarchical closure, and table integrity) to enforce structural compliance and textual accuracy.}
  \label{fig:pipeline}
\end{figure}

We propose \textbf{FireRed-OCR}, a coarse-to-fine progressive training framework designed to transform general-purpose Vision-Language Models (VLMs) into high-precision, structure-aware document understanding experts. As illustrated in Figure~\ref{fig:pipeline}, our pipeline utilizes the Qwen3-VL family as the backbone and proceeds through three distinct stages: \textit{Multi-task Pre-alignment}, \textit{Specialized Supervised Fine-Tuning (SFT)}, and \textit{Format-Constrained Group Relative Policy Optimization (GRPO)}.

\subsection{Stage 1: Multi-task Pre-alignment}
Standard VLMs often hallucinate in OCR tasks involving dense texts or complex layouts. To mitigate this, we first introduce a \textbf{multitask joint training} stage aimed at grounding the model's visual perception in fine-grained textual representations.

Let $\mathcal{D}_{pre}$ be the heterogeneous dataset comprising three distinct sub-tasks:
\begin{itemize}
    \item \textbf{Detection \& OCR ($\mathcal{T}_{det}$):} We train the model to output both bounding boxes and textual content simultaneously. Given an image $I$, the model predicts a sequence $S = \{ (b_i, t_i) \}$, where $b_i$ represents coordinates and $t_i$ is the recognized text. This forces the visual encoder to attend to precise spatial locations.
    \item \textbf{Region OCR ($\mathcal{T}_{reg}$):} To enhance local resolution sensitivity, we employ region-level recognition where the input includes a specific crop or coordinate prompt, and the output is the text within that region.
    \item \textbf{Full-page Markdown ($\mathcal{T}_{md}$):} We introduce initial layout-to-markdown conversion tasks to bridge the modality gap between visual layout and logical structure.
\end{itemize}

The objective function for Stage 1 is the standard auto-regressive cross-entropy loss over the combined dataset:
\begin{equation}
    \mathcal{L}_{Stage1} = - \sum_{(I, y) \in \mathcal{D}_{pre}} \sum_{t=1}^{T} \log P(y_t | y_{<t}, I; \theta_{base})
\end{equation}
This stage yields a model with "Base Document Capability," robust against perceptual errors but potentially lacking in structural formatting consistency.

\subsection{Stage 2: Specialized Supervised Fine-Tuning (SFT)}
Following the pre-alignment, the model possesses basic recognition capabilities but often struggles with the strict formatting required for downstream applications. To bridge this gap, Stage 2 employs \textbf{Specialized SFT} on a curated high-quality dataset $\mathcal{D}_{sft}$, specifically designed to transform raw OCR predictions into structured, machine-readable documents. The training emphasizes four critical dimensions to refine the "Base Document Capability" into a "High-Precision Engine":

\begin{itemize}
    \item \textbf{Structural Consistency:} We explicitly train the model to maintain logical coherence throughout long-context generation. This involves ensuring that the predicted sequence adheres to a valid Document Object Model, preventing fragmentation in lengthy documents.
    \item \textbf{Stability of Hierarchical Expression:} Special attention is given to the precise nesting of document structures. The model learns to strictly distinguish semantic levels, such as headers (e.g., \texttt{\#} vs. \texttt{\#\#}) and nested lists, ensuring the output faithfully reflects the visual hierarchy of the original layout.
    \item \textbf{Standardization of Markdown Formatting:} To mitigate the ambiguity inherent in natural language generation, we enforce a unified Markdown syntax. This involves canonicalizing representations for styling (e.g., bolding, italics) and mathematical expressions, thereby reducing variance in the output space and facilitating easier downstream parsing.
    \item \textbf{Robustness to Cross-lingual and Complex Layouts:} The training distribution is significantly broadened to include diverse linguistic scripts and challenging geometric topologies, such as dense multi-column scientific papers and mixed-media documents. This equips the model with the resilience to handle spatially complex and linguistically heterogeneous content without degradation in recognition accuracy.
\end{itemize}

Through this specialized tuning, the model effectively minimizes the structural hallucinations often observed in general-purpose VLMs, establishing a solid foundation for the subsequent reinforcement learning stage.

\begin{insight}[title=Insight 1: Benefits of Progressive Annotation Refinement.]
We discover that a \textit{coarse-to-fine} data strategy yields superior results compared to using the highest quality data throughout the entire training process. Specifically, training on coarser annotations (PaddleOCR-VL v1) in Stage 1 and finer annotations (v1.5) in Stage 2 achieves higher performance than using the high-precision v1.5 data for both stages.

This implies that coarser data provides a "smoother" learning landscape for the initial training phase, allowing the model to converge on general document understanding capabilities. Introducing the finer annotations later serves as a refinement step, preventing the model from getting stuck in local optima caused by the high complexity of fine-grained labels early in training.
\end{insight}

\subsection{Stage 3: Format-Constrained GRPO}
\label{sec:grpo}
The core innovation of FireRed-OCR is the application of \textbf{Group Relative Policy Optimization (GRPO)}~\cite{shao2024deepseekmath} with strictly defined format constraints. Unlike PPO~\cite{schulman2017proximal} which requires a separate Value model (doubling VRAM usage), GRPO estimates the baseline from the group average of generated outputs, making it highly efficient for high-resolution VLMs~\cite{bai2025qwen3,huang2026step3}.

\paragraph{Group Generation Block.}
For each prompt $x$ (an image-instruction pair), the model $\pi_\theta$ generates a group of $G$ outputs $\{o_1, o_2, ..., o_G\}$ by sampling. These outputs are then evaluated by a set of rule-based reward functions designed to penalize structural syntax errors.

\paragraph{Reward Engineering.}
We design a composite reward function $R(o)$ to guide the model towards structurally valid generation. The total reward is a weighted sum of four components:
\begin{equation}
    R(o) = \lambda_A r_{syntax} + \lambda_B r_{closure} + \lambda_C r_{table} + \lambda_D r_{text}
\end{equation}

\begin{enumerate}
    \item \textbf{Reward A: Formula Syntax ($r_{syntax}$):} We employ a lightweight \LaTeX\ parser/compiler check. If a generated formula string $\mathcal{F}$ fails to compile or contains illegal tokens, $r_{syntax} = -1$; otherwise, it provides a positive score based on complexity.
    \item \textbf{Reward B: Hierarchical Closure ($r_{closure}$):} This reward penalizes unclosed tags. For Markdown/HTML-like structures, or mismatched Markdown tags, we apply a penalty proportional to the number of open nodes. This enforces \textit{structural stability}.
    \item \textbf{Reward C: Table Integrity ($r_{table}$):} Tables are the most fragile structures in OCR. We parse the generated markdown table to check if the number of columns is consistent across all rows. $r_{table} = 1$ if the structure is rectangular and complete, else 0.
    \item \textbf{Reward D: Textual Accuracy ($r_{text}$):} While structure is key, content must remain faithful. We calculate the negative Levenshtein distance (normalized) between the generated text content (stripping structural tags) and the pseudo-ground truth from the SFT model (or human labels if available) to ensure \textit{content correctness}.
\end{enumerate}

\paragraph{Optimization Objective.}
The GRPO objective maximizes the following surrogate objective:
\begin{equation}
    \mathcal{J}_{GRPO}(\theta) = \mathbb{E}_{q \sim P(Q), \{o_i\}_{i=1}^G \sim \pi_{\theta_{old}}(o|q)} \left[ \frac{1}{G} \sum_{i=1}^G \min \left( \frac{\pi_\theta(o_i|q)}{\pi_{\theta_{old}}(o_i|q)} \hat{A}_i, \text{clip}\left(\frac{\pi_\theta(o_i|q)}{\pi_{\theta_{old}}(o_i|q)}, 1-\epsilon, 1+\epsilon\right) \hat{A}_i \right) \right] - \beta \mathbb{D}_{KL}
\end{equation}
where the advantage $\hat{A}_i$ is computed by normalizing the rewards within the group: $\hat{A}_i = \frac{R(o_i) - \text{mean}(\{R(o_j)\})}{\text{std}(\{R(o_j)\})}$.

By optimizing this objective, FireRed-OCR learns to navigate the trade-off between visual grounding and strict syntactic adherence, significantly reducing "Format Collapse" and achieving high generalization on unseen document types.

\begin{insight}[title=Insight 2: Synergistic Gains from Grounding Constraints.]
Interestingly, we observe that applying GRPO-based format constraints \textit{exclusively} to the grounding sub-task leads to a noticeable performance improvement in the main task. This suggests a spillover effect where the structural discipline required for precise coordinate prediction reinforces the model's overall visual-textual alignment. By forcing the model to be rigorous about "where" information is located, we implicitly enhance its ability to understand "what" the information is, thereby benefiting the primary generation task even without direct constraints.
\end{insight}

While the standard pipeline operates sequentially from Stage 1 to Stage 3, we introduce a cyclic refinement mechanism, denoted as \textbf{Iterative SFT-GRPO}. we observe that an iterative application of SFT and GRPO yields superior performance. In this paradigm, the model alternates between Specialized SFT (Stage 2) and Format-Constrained GRPO (Stage 3) for $K$ iterations.

We attribute the success of this iterative strategy to the synergistic interplay between likelihood-based instruction following and reward-based structural enforcement. Specifically:

\begin{enumerate}
    \item \textbf{Decoupling Semantic Fidelity from Structural Rigidity:} 
    FireRed-OCR faces a dual challenge: accurate text recognition (Semantic) and complex layout reconstruction (Structural). SFT primarily ensures the OCR content is correct (preventing hallucinations), while GRPO focuses on strict syntax compliance (e.g., ensuring \LaTeX{} formulas compile or Markdown tables have matching columns). Iterating between them allows the model to refine structural strictness without forgetting the underlying text recognition capabilities.
    
    \item \textbf{Mitigating Reward Hacking in Long-Context Generation:} 
    In pure RL training (GRPO), models may exploit the reward function by generating syntactically perfect but semantically empty or repetitive structures (reward hacking). By periodically re-introducing Specialized SFT using high-quality ground truth, we act as a regularizer that pulls the policy back towards the distribution of meaningful, content-rich document representations.
    
    \item \textbf{Progressive Hard-Constraint Adaptation:} 
    Learning complex constraints (such as hierarchical closure in nested lists or multi-row spanning cells in tables) is difficult in a single pass. The iterative cycle creates a curriculum: early iterations stabilize basic tag generation, while later iterations which guided by GRPO's specific rule-based rewards fine-tune the model to handle corner cases in document layout that standard SFT loss fails to capture.
\end{enumerate}



\section{Experiments}
\label{sec:experiments}

In this section, we present a comprehensive evaluation of FireRed-OCR. We benchmark our model against three categories of state-of-the-art systems: (1) \textbf{General Large Vision-Language Models (VLMs)} such as Qwen3-VL-235B~\cite{bai2025qwen3} and Gemini-3.0 Pro~\cite{google2025gemini3pro}; (2) \textbf{Traditional Pipeline Systems} like PaddleOCR-VL-1.5~\cite{cui2026paddleocr}; and (3) \textbf{Specialized End-to-End (E2E) OCR Models} like DeepSeek-OCR 2~\cite{wei2026deepseek} and dots.ocr~\cite{li2025dots}.

\subsection{Data Preparation}

To develop a robust and versatile model, we constructed a comprehensive training dataset derived from two primary sources: high-quality open-source datasets and re-annotation from open-source datasets. This combination ensures diversity across various document layouts, fonts, and domains.

We integrated a wide range of public datasets to cover specific capabilities such as handwritten text recognition, formula generation, and table parsing.

\textbf{Mathematical Expressions.} 
Handling complex mathematical notation is a core capability of our system. We utilized the \textit{LaTeX OCR} dataset~\cite{latex_ocr} for simple formulas and the \textit{latex-formulas-80M} dataset~\cite{latex-formulas-80M} for complex equations.
A critical challenge addressed during preprocessing was the discrepancy in \LaTeX{} syntax between training sources and standard evaluation benchmarks. We developed a normalization pipeline to unify the \LaTeX{} format, ensuring consistency in spacing, symbol representation, and environment definitions.

\textbf{General Document Understanding and OCR.} 
For general document parsing, we integrated multiple sources.
\begin{itemize}
    \item \textbf{BLIP-3 OCR:} We utilized a subset of the \textit{BLIP-3 OCR} dataset~\cite{xue2024xgen}. Although the original dataset contains 200M samples, we filtered for high-quality bounding box annotations and selected a few samples for training.
    \item \textbf{Docmatix:} Docmatix~\cite{laurencon2024building} provides valuable Visual Question Answering (VQA) data. We re-labeled the OCR ground truth to align with our training objectives. 
\end{itemize}

\textbf{Tabular Data.} 
To enhance the model's ability to interpret structural information, we included the \textit{PubTabNet} dataset~\cite{zhong2019image}. We converted the ground truth from raw HTML representations into a unified Parquet format for efficient loading. 

\textbf{Handwritten Text.} 
To support handwriting recognition, we employed the \textit{IAM Handwriting Database}~\cite{IAM-line}. We processed lines of English handwriting, converting the annotations into Markdown format to maintain consistency with our text generation targets.

\subsection{Implementation Details}
\label{subsec:implementation}

Our training pipeline proceeds in three stages with varying data scales and optimization strategies. 
In Multi-task Pre-alignment stage, we utilize a dataset of approximately 1.3M samples to ground the model's visual perception. 
For pecialized SFT stage, we fine-tune the model on 400k high-quality document-to-markdown pairs to standardize the structural output. 
Both supervised stages employ a global batch size of 256 and a learning rate of $3 \times 10^{-5}$. 
Finally, in the Format-Constrained GRPO stage, we use 50K samples for reinforcement learning and reduce the learning rate to $5 \times 10^{-7}$. 
For generation during the RL phase, we apply nucleus sampling with $p=0.99$ and $k=50$, setting a maximum context length of 24,576 tokens and a completion limit of 2,048 tokens. 
A linear warmup ratio of 0.05 is applied across all training stages.

\subsection{Evaluation}

Our evaluation relies on serveral benchmarks, including OmniDocBench v1.5~\cite{ouyang2025omnidocbench}, OCRBench~\cite{liu2024ocrbench}, TEDS, PubTabNet~\cite{zhong2019image} and FireRedBench. FireRedBench is a newly introduced internal benchmark focusing on complex, non-standard layouts (e.g., distorted scans, dense multi-column papers, and embedded logic diagrams). This dataset is designed to stress-test layout robustness where traditional crop-based methods often fail.

We report \textbf{Overall} scores alongside specific metrics: \textbf{Edit Distance} (Text/Reading Order, lower is better), \textbf{CDM} (Formula match), and \textbf{TEDS} (Table structure).

\subsection{Main Results}

\begin{table}[h]
    \centering
    \caption{\textbf{Overall Comparison.} FireRed-OCR-2B outperforms massive general VLMs and competitive pipelines, particularly balancing standard accuracy with in-the-wild robustness. For PaddleOCR-VL-1.5 and GLM-OCR on OCRBench (Text), scores are reported as API / pure VLM.}
    \label{tab:overall}
    \resizebox{0.9\textwidth}{!}{
    \begin{tabular}{l c c c c c}
        \toprule
        \textbf{Model} & \textbf{OmniDocBench v1.5} & \textbf{FireRedBench} & \textbf{OCRBench (Text)} & \textbf{TEDS} & \textbf{PubTabNet} \\
        \midrule
        \multicolumn{6}{l}{\textit{Pipeline Systems}} \\
        MinerU2.5~\cite{niu2025mineru2} & 90.67 & - & 75.3 & 85.4 & 88.4 \\
    PaddleOCR-VL-1.5~\cite{cui2026paddleocr} & 94.50 & 76.47 & 53.5 / 87.0 & 83.3 & 84.6 \\
        GLM-OCR~\cite{glmocr} & 94.60 & 74.33 & 
      61.0 / 95.0 & 86.0 & 85.2 \\
        \midrule
        \multicolumn{6}{l}{\textit{General VLMs }} \\
        GPT-5.2~\cite{gpt52} \faLock & 85.50 & - & 83.7 & 67.6 & 84.4 \\
        Gemini-3.0 Pro~\cite{google2025gemini3pro} \faLock & 90.33 & 79.68 & 91.9 & 81.8 & 91.4 \\
        Qwen3-VL-235B-A22B~\cite{bai2025qwen3} & 89.15 & 79.04 & \textbf{95.0} & - & - \\
        Qwen3.5-397B-A17B~\cite{qwen3.5} & 90.80 & 81.85 & 90.5 &64.9 & 70.3\\  
        \midrule
        \multicolumn{6}{l}{\textit{ End-to-End OCR Models}} \\
        dots.ocr~\cite{li2025dots} & 88.41 & 72.93 & 92.1 & 62.4 & 71.0 \\
        DeepSeek-OCR 2~\cite{wei2026deepseek} & 91.09 & 61.61 & 48.5 & - & - \\
     
        \rowcolor{red!10}\textbf{FireRed-OCR-2B (Ours)} & \textbf{92.94} & \textbf{74.62} & \textbf{93.5} & \textbf{80.6} & \textbf{77.0} \\
        
        \bottomrule
    \end{tabular}
    }
\end{table}

\subsubsection{Overall Performance Comparison}
As presented in Table~\ref{tab:overall}, we conduct a comprehensive benchmarking of \textbf{FireRed-OCR-2B} against three distinct categories of state-of-the-art methods: Pipeline Systems, massive General Vision-Language Models (VLMs), and specialized End-to-End (E2E) models. Despite its compact size, our model demonstrates remarkable performance, balancing standard accuracy with in-the-wild robustness.

\paragraph{Comparison with End-to-End Specialists.}
Within the domain of specialized E2E models, FireRed-OCR-2B establishes a new state-of-the-art. On the challenging \textit{OmniDoc v1.5} benchmark, our model achieves a score of \textbf{92.94}, surpassing competitive baselines such as DeepSeek-OCR 2 (91.09) and dots.ocr (88.41) by a significant margin. Similarly, on \textit{OCRBench (Text)}, we achieve a score of \textbf{93.5}, outperforming the nearest E2E competitor by over 1.4 points. These results underscore the efficacy of our proposed architecture and training strategy in handling complex document understanding tasks without relying on external detection modules.

\paragraph{Efficiency against General VLMs.}
A critical observation is the parameter efficiency of FireRed-OCR-2B. While general VLMs like Qwen3-VL-235B and Gemini-3.0 Pro utilize hundreds of billions of parameters, our 2B-parameter model delivers highly competitive results. Notably, on \textit{OCRBench}, FireRed-OCR-2B (93.5) outperforms GPT-5.2 (83.7) and Gemini-3.0 Pro (91.9), and is comparable to the massive Qwen3-VL-235B (95.0). Furthermore, on \textit{OmniDoc v1.5}, our model surpasses Qwen3-VL-235B (89.15) and Qwen3.5-397B (90.80), demonstrating that a well-optimized, domain-specific model can outperform substantially larger generalist models in fine-grained optical character recognition tasks.

\paragraph{Robustness compared to Pipeline Systems.}
Table~\ref{tab:overall} reveals that FireRed-OCR-2B challenges the dominance of heavy pipeline systems. Despite lacking the specialized layout analysis modules found in systems like MinerU2.5, our 2B-parameter model delivers remarkable results on text-centric benchmarks. Specifically, on \textit{OmniDocBench v1.5}, we surpass MinerU2.5 (92.94 vs. 90.67) and approach the performance of GLM-OCR (94.60). Furthermore, on \textit{OCRBench}, our single-model architecture (93.5) significantly outperforms the API-based configurations of pipeline competitors and remains competitive with their pure VLM counterparts, demonstrating that massive ensembles are not strictly necessary for high-accuracy text recognition.

\subsubsection{Detailed Analysis on OmniDocBench v1.5}
Table~\ref{tab:omnidoc} presents a granular analysis of model performance across diverse document elements, including text, formulas, tables, and reading order reconstruction. \textbf{FireRed-OCR-2B} demonstrates exceptional capabilities, achieving an Overall score of \textbf{92.94}. This result not only establishes a new state-of-the-art among End-to-End (E2E) models but also challenges the dominance of massive General VLMs and complex Pipeline Systems.

\paragraph{Defying Scaling Laws against General VLMs.}
A striking observation is that FireRed-OCR-2B significantly outperforms general-purpose models with vastly larger parameter counts. Our 2B model surpasses \textbf{Qwen3.5-397B} (90.80) and \textbf{Gemini-3.0 Pro} (90.33), despite being less than 1\% of their size. In specific tasks such as table structure recognition, our model achieves a $\text{Table}^{\text{TEDS}}$ score of 90.31, dramatically outperforming Qwen3-VL-235B (86.21) and GPT-4o (67.07). This suggests that domain-specific architectural optimization is far more parameter-efficient than brute-force scaling for document understanding tasks.

\paragraph{Dominance in End-to-End OCR.}
Within the E2E category, FireRed-OCR-2B outperforms the previous best model, DeepSeek-OCR 2 (91.09), across all metrics. Crucially, we conduct a direct comparison with \textbf{Qwen3-VL-2B}, a general VLM of equivalent size (separated by the dashed line in Table~\ref{tab:omnidoc_detailed}). As highlighted in the bottom row ($\Delta$), our model delivers a massive performance leap: an \textbf{+11.07} point improvement in Overall score and a \textbf{+20.54} point increase in Table TEDS. This validates the effectiveness of our training recipe over standard VLM pre-training for OCR tasks.

\paragraph{Precision in Text and Layout.}
Our model exhibits pixel-level precision. In raw text recognition, FireRed-OCR-2B achieves the lowest edit distance ($\text{Text}^{\text{Edit}}$) of \textbf{0.032} among all evaluated models, surpassing even the top-performing pipeline system, PaddleOCR-VL-1.5 (0.035). Furthermore, our Reading Order score ($\text{R-order}^{\text{Edit}}$) of 0.041 is the best in the table, indicating superior layout comprehension logic. While Pipeline Systems like PaddleOCR-VL-1.5 maintain a slight edge in complex formulas ($\text{Formula}^{\text{CDM}}$), our E2E approach (91.71) significantly narrows the gap compared to other E2E baselines, offering a more balanced trade-off between accuracy and system simplicity.

\begin{table}[t]
    \centering
    \caption{\textbf{Detailed Evaluation on OmniDocBench v1.5.}.}
    \label{tab:omnidoc}
    \resizebox{\textwidth}{!}{
    \begin{tabular}{lcccccc}
        \toprule
        \textbf{Model} & \textbf{Overall} $\uparrow$ & \textbf{Text}$^{\text{Edit}}$ $\downarrow$ & \textbf{Formula}$^{\text{CDM}}$ $\uparrow$ & \textbf{Table}$^{\text{TEDS}}$ $\uparrow$ & \textbf{Table}$^{\text{TEDS\_s}}$ $\uparrow$ & \textbf{R-order}$^{\text{Edit}}$ $\downarrow$ \\
        \midrule
       \multicolumn{7}{l}{\textit{Pipeline OCR Systems}} \\
        Marker-1.8.2~\cite{markerpdf} & 71.30 & 0.206 & 76.66 & 57.88 & 71.17 & 0.250 \\
        MinerU2-pp~\cite{wang2024mineru} & 71.51 & 0.209 & 76.55 & 70.90 & 79.11 & 0.225 \\
        Dolphin~\cite{feng2025dolphin} & 74.67 & 0.125 & 67.85 & 68.70 & 77.77 & 0.124 \\
        Dolphin-1.5~\cite{dolphin2025} & 83.21 & 0.092 & 80.78 & 78.06 & 84.10 & 0.080 \\
        PP-StructureV3~\cite{cui2025paddleocr} & 86.73 & 0.073 & 85.79 & 81.68 & 89.48 & 0.073 \\
        MonkeyOCR-pro-1.2B~\cite{li2025monkeyocr} & 86.96 & 0.084 & 85.02 & 84.24 & 89.02 & 0.130 \\
        MonkeyOCR-3B~\cite{li2025monkeyocr} & 87.13 & 0.075 & 87.45 & 81.39 & 85.92 & 0.129 \\
        MonkeyOCR-pro-3B~\cite{li2025monkeyocr} & 88.85 & 0.075 & 87.25 & 86.78 & 90.63 & 0.128 \\
        MinerU2.5~\cite{niu2025mineru2} & 90.67 & 0.047 & 88.46 & 88.22 & 92.38 & 0.044 \\
        PaddleOCR-VL~\cite{cui2025paddleocr2} & 92.86 & 0.035 & 91.22 & 90.89 & 94.76 & 0.043 \\
        \textbf{PaddleOCR-VL-1.5}~\cite{cui2026paddleocr} & \textbf{94.50} & 0.035 & \textbf{94.21} & \textbf{92.76} & \textbf{95.79} & 0.042 \\
        \midrule
        \multicolumn{7}{l}{\textit{General VLMs}} \\
        GPT-4o~\cite{hurst2024gpt} \faLock & 75.02 & 0.217 & 79.70 & 67.07 & 76.09 & 0.148 \\
        GPT-5.2~\cite{gpt52} \faLock & 85.50 & 0.123 & 86.11 & 82.66 & 87.35 & 0.099 \\
        Gemini-2.5 Pro~\cite{google2025gemini25pro} \faLock & 88.03 & 0.075 & 85.82 & 85.71 & 90.29 & 0.097 \\
        Gemini-3.0 Pro~\cite{google2025gemini3pro} \faLock & 90.33 & 0.065 & 89.18 & 88.28 & 90.29 & 0.071 \\
        
        InternVL3-76B~\cite{zhu2025internvl3} & 80.33 & 0.131 & 83.42 & 70.64 & 77.74 & 0.113 \\
        Qwen2.5-VL-72B~\cite{Qwen2.5-VL} & 87.02 & 0.094 & 88.27 & 82.15 & 86.22 & 0.102 \\
        InternVL3.5-241B~\cite{wang2025internvl3} & 82.67 & 0.142 & 87.23 & 75.00 & 81.28 & 0.125 \\
        Qwen3-VL-235B-A22B~\cite{bai2025qwen3} & 89.15 & 0.069 & 88.14 & 86.21 & 90.55 & 0.068 \\
        Qwen3.5-397B-A17B~\cite{qwen3.5} & 90.80 & - & -&- & -\\
        
        \midrule
        \multicolumn{7}{l}{\textit{End-to-End OCR Models}} \\
        OCRFlux-3B~\cite{OCRFlux} & 74.82 & 0.193 & 68.03 & 75.75 & 80.23 & 0.202 \\
        Mistral OCR~\cite{mistral-ocr} & 74.82 & 0.193 & 68.03 & 75.75 & 80.23 & 0.202 \\
        POINTS-Reader~\cite{liu2025points} & 80.98 & 0.134 & 79.20 & 77.13 & 81.66 & 0.145 \\
        olmOCR-7B~\cite{olmocrbench} & 81.79 & 0.096 & 86.04 & 68.92 & 74.77 & 0.121 \\
        MinerU2-VLM~\cite{wang2024mineru} & 85.56 & 0.078 & 80.95 & 83.54 & 87.66 & 0.086 \\
        Nanonets-OCR-s~\cite{Nanonets-OCR-S} & 85.59 & 0.093 & 85.90 & 80.14 & 85.57 & 0.108 \\
        dots.ocr~\cite{li2025dots} & 88.41 & 0.048 & 83.22 & 86.78 & 90.62 & 0.053 \\
        OCRVerse~\cite{zhong2026ocrverse} & 88.56 & 0.058 & 86.91 & 84.55 & 88.45 & 0.071 \\
        DeepSeek-OCR~\cite{wei2025deepseek} & 87.36 & 0.073 & 84.14 & 85.25 & 89.01 & 0.085 \\
        DeepSeek-OCR 2~\cite{wei2026deepseek} & 91.09 & 0.048 & 90.31 & 87.75 & 92.06 & 0.057 \\
        \hdashline
        Qwen3-VL-2B~\cite{bai2025qwen3} & 81.87 & 0.100 & 85.87 & 69.77 & 74.37 & 0.115 \\
        \rowcolor{red!10}\textbf{FireRed-OCR-2B (Ours)} & \textbf{92.94} & \textbf{0.032} & \textbf{91.71} & \textbf{90.31} & \textbf{93.81} & \textbf{0.041} \\
        $\Delta$ & +11.07 & +0.068 & +5.84 & +20.54 & +19.44 & +0.074 \\
       
        \bottomrule
    \end{tabular}
    }
\end{table}

\subsubsection{Performance on FireRedBench}
Table~\ref{tab:firered} details the evaluation results on \textit{FireRedBench}, a challenging benchmark specifically designed to test model robustness against complex document layouts and diverse topologies. In this demanding scenario, \textbf{FireRed-OCR-2B} achieves an Overall score of \textbf{74.62}.

\paragraph{Surpassing Pipeline Complexity.}
A notable finding is that our compact E2E model outperforms the complex \textit{GLM-OCR} pipeline (74.62 vs. 74.33) and remains highly competitive with \textit{PaddleOCR-VL-1.5} (76.47). While pipeline systems typically rely on separate detection and recognition modules to handle layout variability, FireRed-OCR-2B effectively internalizes these capabilities within a single 2B-parameter architecture. This suggests that our model learns a more robust internal representation of document structure, avoiding the cascading errors often observed in multi-stage pipelines when processing intricate layouts.

\paragraph{State-of-the-Art among E2E Models.}
Within the realm of End-to-End models, FireRed-OCR-2B establishes a significant lead. It outperforms \textit{DeepSeek-OCR 2} by a remarkable margin (+13.01 points overall) and surpasses \textit{dots.ocr} (72.93). Specifically, in structure-heavy tasks, our model achieves a $\text{Table}^{\text{TEDS}}$ of 65.63, substantially higher than DeepSeek-OCR 2 (55.06), highlighting its proficiency in parsing dense tabular data without external layout analysis tools.

\paragraph{Significant Gains over Base VLM.}
To isolate the contribution of our training recipe, we compare FireRed-OCR-2B directly with its backbone, \textit{Qwen3-VL-2B} (separated by the dashed line). As shown in the $\Delta$ row, our method yields a comprehensive improvement across all metrics, with an Overall increase of \textbf{+9.04}. Most strikingly, we observe a \textbf{+15.78} point jump in $\text{Table}^{\text{TEDS}}$ and a \textbf{+16.64} point increase in $\text{Table}^{\text{TEDS\_s}}$, confirming that our supervised fine-tuning strategy fundamentally enhances the model's ability to ground visual structure into textual representations.

\begin{table}[h]
    \centering
    \caption{\textbf{Evaluation on FireRedBench.} This benchmark tests robustness against complex layouts. Note the significant performance drop of Pipeline models compared to the stability of E2E models.}
    \label{tab:firered}
    \resizebox{\textwidth}{!}{
    \begin{tabular}{lcccccc}
        \toprule
        \textbf{Model} & \textbf{Overall} $\uparrow$ & \textbf{Text}$^{\text{Edit}}$ $\downarrow$ & \textbf{Formula}$^{\text{CDM}}$ $\uparrow$ & \textbf{Table}$^{\text{TEDS}}$ $\uparrow$ & \textbf{Table}$^{\text{TEDS\_s}}$ $\uparrow$ & \textbf{R-order}$^{\text{Edit}}$ $\downarrow$ \\
        \midrule
        \multicolumn{7}{l}{\textit{Pipeline OCR Systems}} \\
        GLM-OCR~\cite{glmocr} & 74.33 & 0.309 & 82.53 & 71.35 & 79.93 & 0.456 \\
        PaddleOCR-VL-1.5~\cite{cui2026paddleocr} & 76.47 & 0.291 & 92.37 & 66.15 & 74.39 & 0.453 \\
        \midrule
        \multicolumn{7}{l}{\textit{General VLMs}} \\
        GPT-5.2~\cite{gpt52} \faLock & 68.09 & 0.238 & 66.33 & 61.74 & 68.00 & 0.380 \\ 
        Gemini-3.0 Pro~\cite{google2025gemini3pro} \faLock & 79.68 & 0.169 & 80.11 & 75.82 & 82.73 & 0.353 \\
        Qwen3-VL-235B~\cite{bai2025qwen3} & 79.04 & 0.208 & 90.83 & 67.09 & 74.52 & 0.366 \\
        Qwen3.5-397B-A17B~\cite{qwen3.5} & 81.85 & 0.205 & 94.14 & 71.93 & 78.25 & 0.383 \\

        \midrule
        \multicolumn{7}{l}{\textit{End-to-End OCR Models}} \\
        DeepSeek-OCR 2~\cite{wei2026deepseek} & 61.61 & 0.290 & 58.78 & 55.06 & 59.42 & 0.437 \\
        dots.ocr~\cite{li2025dots} & 72.93 & 0.240 & 82.53 & 60.25 & 64.08 & 0.419 \\
        \hdashline
        Qwen3-VL-2B~\cite{bai2025qwen3} & 65.58 & 0.283 & 75.19 & 49.85 & 55.66 & 0.388 \\
        \rowcolor{red!10}\textbf{FireRed-OCR-2B (Ours)} & 74.62 & 0.248 & 83.02 & 65.63 & 72.30 & 0.430 \\
        $\Delta$ & +9.04 & +0.035 & +7.83 & +15.78 & +16.64 & -0.042 \\
       
        \bottomrule
    \end{tabular}
    }
\end{table}

\subsection{Ablation Study}

In this section, we investigate the impact of different data distributions and training strategies on the model's performance. We employ Group Relative Policy Optimization (GRPO) across three primary domains: Text, Table, and Formula (LaTeX). The results are summarized in Table~\ref{tab:ablation}.

\subsubsection{Single-Domain Optimization}
As shown in the first block of Table~\ref{tab:ablation}, applying GRPO to specific domains yields significant improvements in their respective metrics. Specifically, the \textit{Base + Table GRPO} configuration achieves a substantial gain in Table TEDS (+2.0 points), while \textit{Base + Text GRPO} minimizes the Text Edit distance to 0.04. This demonstrates that GRPO effectively aligns the model with domain-specific rewards. However, we observe that single-domain optimization often fails to generalize well to other modalities (e.g., Table GRPO underperforms in Text Edit compared to the Text-only baseline).

\subsubsection{The Challenge of Modality Interference}
A naive combination of datasets does not guarantee performance improvements. The \textit{Base + Table + Text} experiment yields an Overall score of 88.72, which is inferior to the pure text baseline (88.91) and the single-domain Table GRPO (89.18). We hypothesize that this sub-optimal performance stems from modality interference, where the optimization direction of one modality conflicts with another during the joint alignment process, leading to a "alignment tax" that degrades general performance.

\subsubsection{Balanced Mixture Strategy}
To address the interference issue, we propose a balanced sampling strategy. As evidenced in the last row of Table~\ref{tab:ablation}, the \textit{Base + Mix 1:1:1} configuration achieves the state-of-the-art Overall score of \textbf{89.60}. Notably, this setting outperforms the \textit{Table GRPO (Extended)} setting (89.28). This indicates that a balanced distribution of multimodal supervision signals rather than simple data accumulation, which is critical for mitigating modality competition and maximizing the comprehensive capability of the model.

\begin{table}[h]
    \centering
    \caption{\textbf{Ablation study on GRPO strategies.}}
    \label{tab:ablation}
    \resizebox{\textwidth}{!}{
    \begin{tabular}{lcccccc}
        \toprule
        \textbf{Model} & \textbf{Overall} $\uparrow$ & \textbf{Text}$^{\text{Edit}}$ $\downarrow$ & \textbf{Formula}$^{\text{CDM}}$ $\uparrow$ & \textbf{Table}$^{\text{TEDS}}$ $\uparrow$ & \textbf{Table}$^{\text{TEDS\_s}}$ $\uparrow$ & \textbf{R-order}$^{\text{Edit}}$ $\downarrow$ \\
        \midrule
        \textit{Baselines} & & & & &\\
        Base & 88.38 & 0.047 & 87.72 & 82.14 & 85.63 & 0.060\\
        \midrule
        \textit{Single-Domain GRPO} & & & & &\\
        Base + Table GRPO & 89.18 & 0.046 & 88.01  & 84.14  & 88.08 & 0.059\\
        Base + Text GRPO & 88.91 & 0.040 & 88.16 & 82.57 & 86.14 & 0.049\\
        Base + LaTeX GRPO & 88.86 & 0.049 & 87.57 & 83.90 & 87.42 & 0.061 \\
        \midrule
        \textit{Joint Training \& Mixture} & & & & &\\
        Base + Table + Text (Joint) & 88.72 & 0.041 & 86.95  & 83.31 & 87.06 & 0.050\\
        Base + Table GRPO (Extended) & 89.28 & 0.047 & 87.01 & 85.54 & 89.44 & 0.062 \\
        \textbf{Base + Mix 1:1:1} &  89.60  &  0.035  & 86.24 & \ 86.07  & 89.73 & 0.047\\
        \bottomrule
    \end{tabular}
    }
\end{table}

\subsection{Qualitative Evaluation}
We present a qualitative evaluation of our model across four distinct domains: mathematical expression parsing, handwritten text recognition, complex document layout analysis, and structural table reconstruction. The following cases demonstrate the model's capability to bridge the gap between raw pixels and structured semantic understanding.

\subsubsection{Case 1: Mathematical Structure Parsing}
Traditional OCR fails at 2D spatial structures like limits and nested fractions. Our model successfully translates the pixel information into a syntactically correct \LaTeX{} or Markdown sequence, preserving the logical derivation of the \textit{Product Rule}.
\vspace{1cm}
\begin{figure}[hbt!]
    \centering
    
    \begin{subfigure}[b]{0.48\textwidth}
        \centering
        \includegraphics[width=\linewidth]{figures/case/Z1.png} 
        \caption{\textbf{Input:} Original Slide Image}
    \end{subfigure}
    \hfill 
    \begin{subfigure}[b]{0.48\textwidth}
        \centering
        \includegraphics[width=\linewidth]{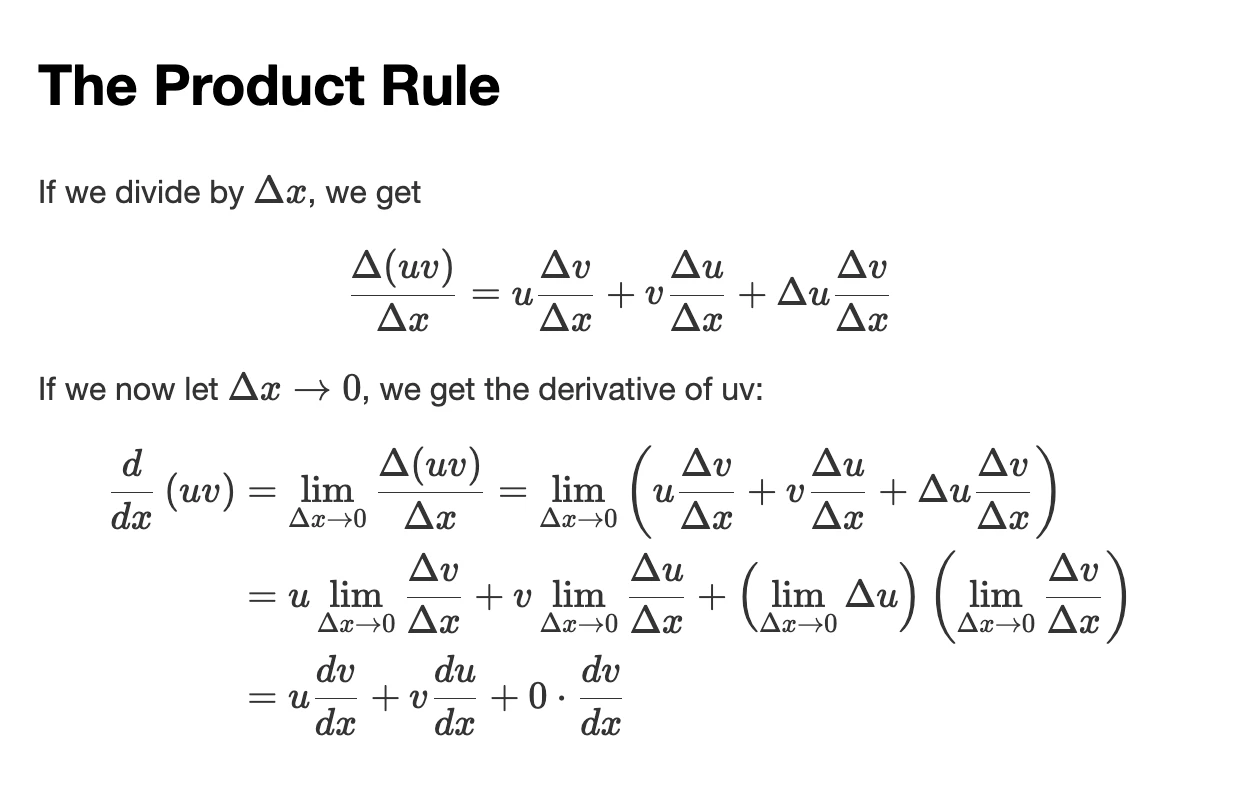} 
        \caption{\textbf{Output:} Rendered \LaTeX{} Result}
    \end{subfigure}
    \caption{\textbf{From Pixels to Formula:} The model accurately identifies complex structures such as $\lim_{\Delta x\to0}$ and fraction hierarchies, enabling direct digitization of STEM materials.}
    \label{fig:case1}
\end{figure}

\vspace{12cm}

\subsubsection{Case 2: Robust Handwritten Text Recognition}
Dealing with unconstrained handwriting on noisy backgrounds (e.g., lined paper) is challenging. The model demonstrates high robustness by effectively separating foreground text from background lines and recognizing cursive strokes.

\begin{figure}[hbt!]
    \centering
    \begin{subfigure}[b]{0.48\textwidth}
        \centering
        \includegraphics[width=\linewidth]{figures/case/A1.jpeg} 
        \caption{\textbf{Input:} Handwritten Essay}
    \end{subfigure}
    \hfill
    \begin{subfigure}[b]{0.48\textwidth}
        \centering
        \includegraphics[width=\linewidth]{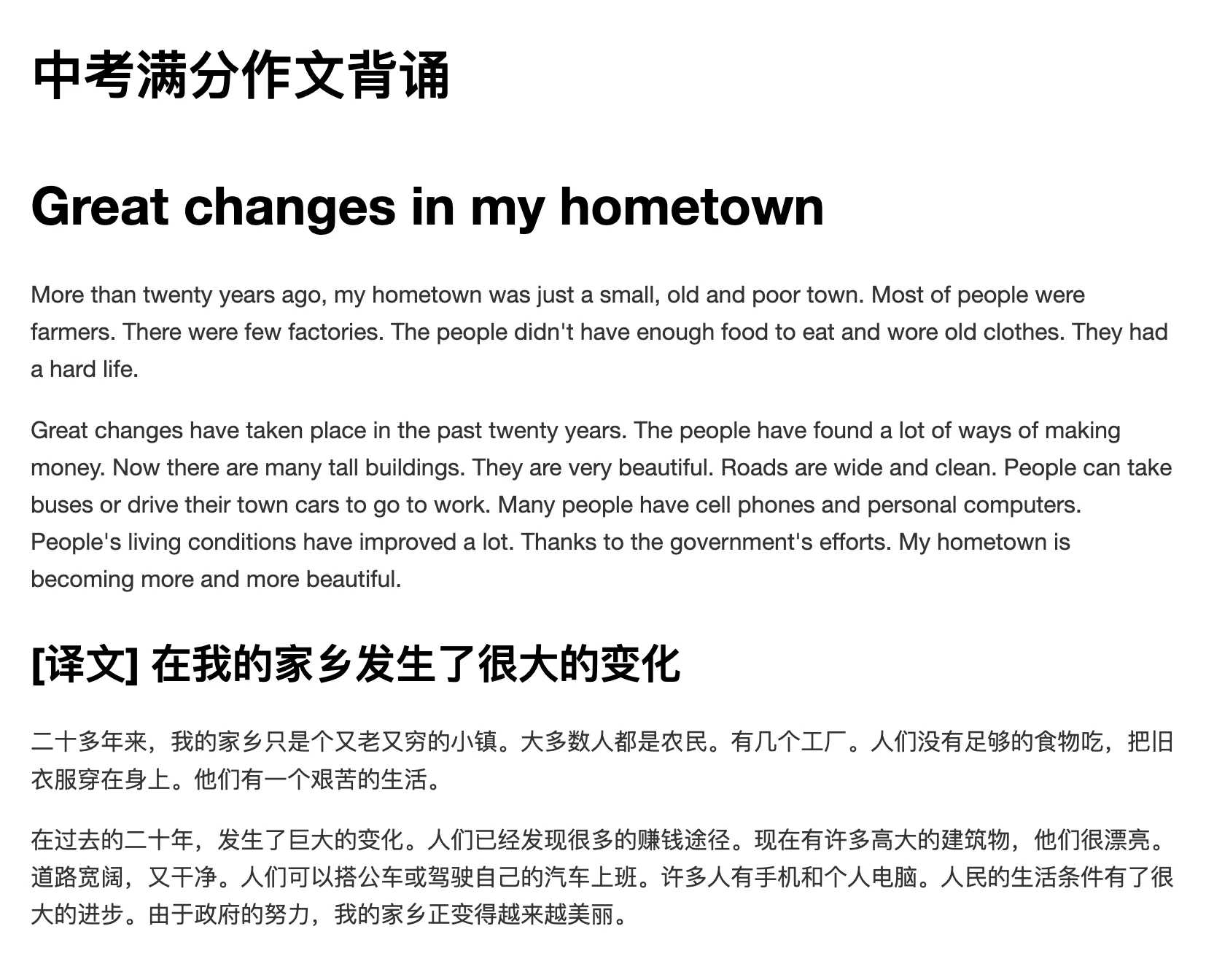} 
        \caption{\textbf{Output:} Digitized Text}
    \end{subfigure}
    \caption{\textbf{Unconstrained Handwriting:} Despite the interference of grid lines and personal writing styles, the model achieves near-perfect recall on the essay content.}
    \label{fig:case2}
\end{figure}

\clearpage 
\subsubsection{Case 3: Complex Layout Analysis (Newspaper)}
Newspapers often feature mixed orientations (vertical/horizontal text) and multi-column layouts. The model correctly performs Layout Analysis (LA) to recover the logical Reading Order (RO), handling vertical CJK text correctly.

\begin{figure}[hbt!]
    \centering
    \begin{subfigure}[b]{0.48\textwidth}
        \centering
        \includegraphics[width=\linewidth]{figures/case/J1.jpg} 
        \caption{\textbf{Input:} Complex Newspaper Layout}
    \end{subfigure}
    \hfill
    \begin{subfigure}[b]{0.48\textwidth}
        \centering
        \includegraphics[width=0.4\linewidth]{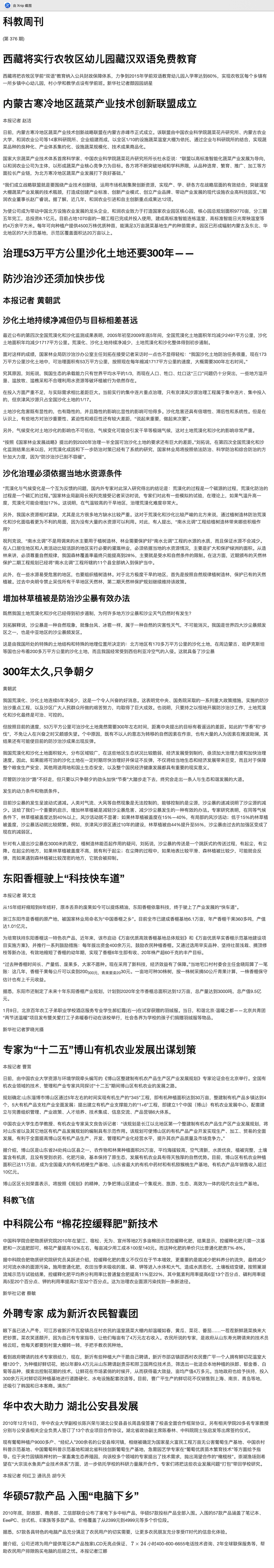} 
        \caption{\textbf{Output:} Reconstructed Layout}
    \end{subfigure}
    \caption{\textbf{Layout \& Reading Order:} The model distinguishes between the main article and sidebars, correctly parsing vertical text flows common in Chinese publications. 
    }
    \label{fig:case3}
\end{figure}

\subsubsection{Case 4: Structural Table Reconstruction}
\textbf{Insight:} Financial tables contain dense information with spanning cells. The model reconstructs the logical tree of the table, correctly identifying row spans (e.g., grouping stocks by category) and hierarchical headers.

\begin{figure}[hbt!]
    \centering
    \begin{subfigure}[b]{0.48\textwidth}
        \centering
        \includegraphics[width=\linewidth]{figures/case/C1.png} 
        \caption{\textbf{Input:} Financial Report Image}
    \end{subfigure}
    \hfill
    \begin{subfigure}[b]{0.48\textwidth}
        \centering
        \includegraphics[width=\linewidth]{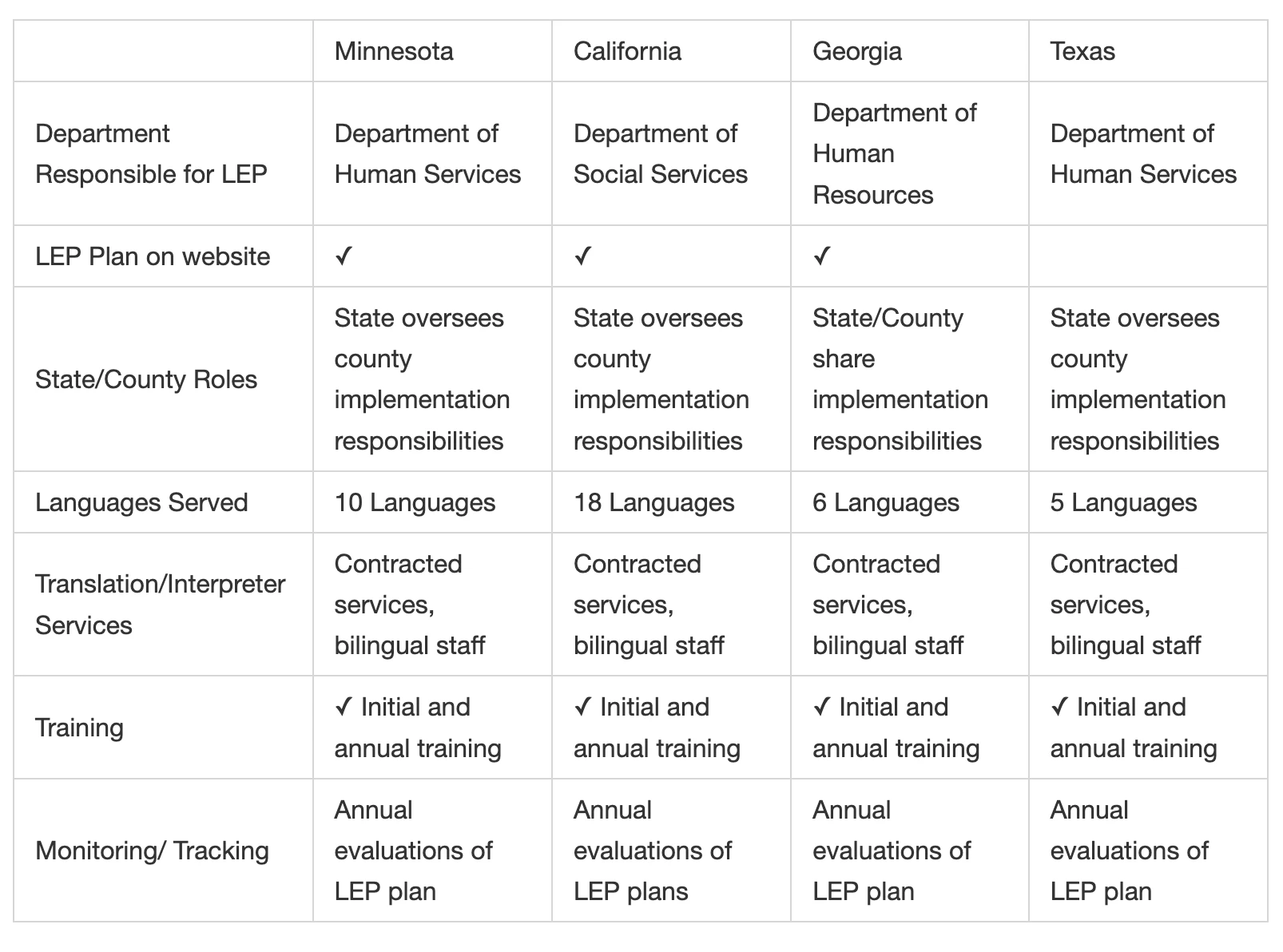} 
        \caption{\textbf{Output:} Rendered Result}
    \end{subfigure}
    \caption{\textbf{Table Structure Recovery}.}
    \label{fig:case4}
\end{figure}

\begin{figure}[hbt!]
    \centering
    \begin{subfigure}[b]{0.48\textwidth}
        \centering
        \includegraphics[width=\linewidth]{figures/case/F1.png} 
        \caption{\textbf{Input:} Financial Report Image}
    \end{subfigure}
    \hfill
    \begin{subfigure}[b]{0.48\textwidth}
        \centering
        \includegraphics[width=\linewidth]{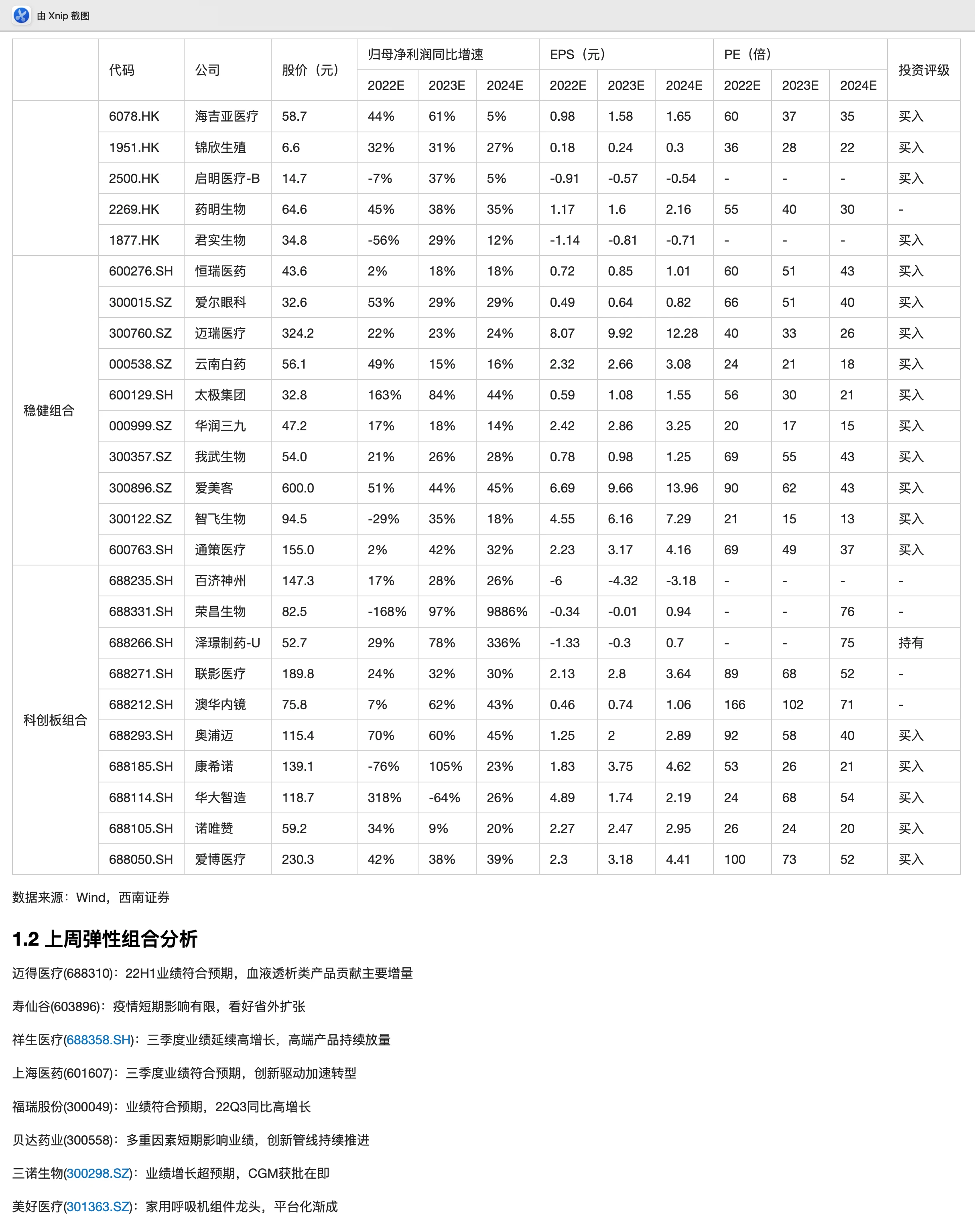} 
        \caption{\textbf{Output:} Rendered Result}
    \end{subfigure}
    \caption{\textbf{Table Structure Recovery}.}
    \label{fig:case4}
\end{figure}

\section{Conclusion}
In this work, we presented FireRed-OCR, an open-source document parsing framework that addresses the critical issue of structural hallucination in VLMs. By implementing a ``Geometry + Semantics'' data engine and a ``Three-Stage Progressive Training'' strategy, we successfully evolved a general purpose VLM (Qwen3-VL) into a pixel-precise structural expert. Our results on OmniDocBench v1.5 confirm that even 2B parameter models can achieve SOTA performance with high-quality data and specialized training constraints. We hope this work provides a reproducible paradigm for the community to transition from general multimodal models to specialized structured document models.

\section{Authors}
\begin{itemize}
   \item \textbf{Core Contributors:} Hao Wu, Haoran Lou, Xinyue Li, Zuodong Zhong, Zhaojun Sun, Phellon Chen, Xuanhe Zhou, Kai Zuo, Yibo Chen, Xu Tang, Yao Hu
    \item \textbf{Contributors:} Boxiang Zhou, Jian Wu, Yongji Wu, Wenxin Yu, Yingmiao Liu, Yuhao Huang, Manjie Xu, Gang Liu, Yidong Ma, Zhichao Sun, Changhao Qiao
\end{itemize}

\clearpage

\addcontentsline{toc}{section}{References}
\bibliography{redocr_report}
\bibliographystyle{plain}


\end{document}